\documentclass[conference]{IEEEtran}
\IEEEoverridecommandlockouts
\usepackage{cite}
\usepackage{graphicx}
\usepackage{amsmath,amssymb,amsfonts}
\usepackage{algorithmic}
\usepackage{graphicx}
\usepackage{textcomp}
\usepackage{xcolor}
\usepackage[numbers,sort&compress]{natbib}
\usepackage{graphicx}
\usepackage{threeparttable}
\usepackage{multirow}
\usepackage{lmodern}
\usepackage{tcolorbox}
\usepackage{caption}
\usepackage{booktabs}
\usepackage{lipsum}
\usepackage{tabularx} 
\usepackage{adjustbox}
\usepackage{etoolbox}

\newcommand{\teblebold}[1]{\footnotesize\textbf{#1}}

\def\BibTeX{{\rm B\kern-.05em{\sc i\kern-.025em b}\kern-.08em
    T\kern-.1667em\lower.7ex\hbox{E}\kern-.125emX}}
\begin{document}

\title{Pyramid Transformer for Traffic Sign Detection\\
}


\makeatletter
\newcommand{\linebreakand}{%
  \end{@IEEEauthorhalign}
  \hfill\mbox{}\par
  \mbox{}\hfill\begin{@IEEEauthorhalign}
}
\makeatother

\author{\IEEEauthorblockN{Omid Nejati Manzari}
\IEEEauthorblockA{\textit{School of Electrical Engineering,} \\
\textit{Iran University of Science and Technology,}\\
Tehran, Iran \\
omid\_nejaty@elec.iust.ac.ir}
\and
\IEEEauthorblockN{Amin Boudesh}
\IEEEauthorblockA{\textit{Department of Mechanical Engineering,} \\
\textit{Tarbiat Modares Univesity,}\\
Tehran, Iran \\
Amin.bdsh@modares.ac.ir}
\linebreakand
\IEEEauthorblockN{Shahriar B. Shokouhi}
\IEEEauthorblockA{\textit{School of Electrical Engineering,} \\
\textit{Iran University of Science and Technology,}\\
Tehran, Iran \\
bshokouhi@iust.ac.ir}
}

\maketitle

\begin{abstract}
Traffic sign detection is a vital task in the visual system of self-driving cars and the automated driving system. Recently, novel Transformer-based models have achieved encouraging results for various computer vision tasks. We still observed that vanilla ViT could not yield satisfactory results in traffic sign detection because the overall size of the datasets is very small and the class distribution of traffic signs is extremely unbalanced. To overcome this problem, a novel Pyramid Transformer with locality mechanisms is proposed in this paper. Specifically, Pyramid Transformer has several spatial pyramid reduction layers to shrink and embed the input image into tokens with rich multi-scale context by using atrous convolutions. Moreover, it inherits an intrinsic scale invariance inductive bias and is able to learn local feature representation for objects at various scales, thereby enhancing the network robustness against the size discrepancy of traffic signs. The experiments are conducted on the German Traffic Sign Detection Benchmark (GTSDB). The results demonstrate the superiority of the proposed model in the traffic sign detection tasks. More specifically, Pyramid Transformer achieves $77.8\%$ $mAP$ on GTSDB when applied to the Cascade RCNN as the backbone, which surpasses most well-known and widely-used state-of-the-art models.
\end{abstract}


\begin{figure*}
	\centering
	\includegraphics[width=\textwidth]{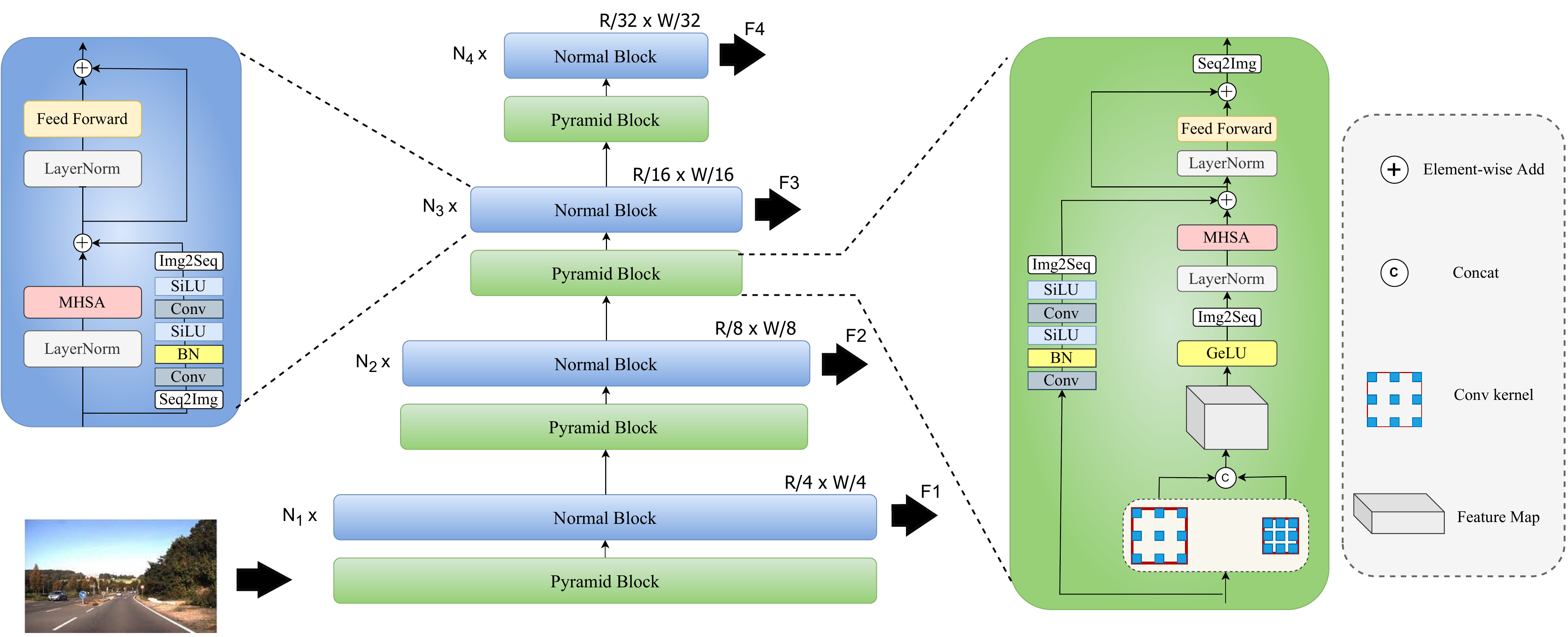}
    \caption{\textbf{Network architecture of the proposed Pyramid Transformer.}}
	\label{FIG:2}
\end{figure*}

\section{Introduction}

Traffic sign recognition and detection (TSRD) are crucial to driver assistance systems and self-driving cars. One of the main prerequisites for the widespread implementation of TSRD is an algorithm that is robust, reliable, and high accuracy in various real-world scenarios. Besides the large variation in traffic signs to detect, the traffic images taken on the road are not ideal. They are often distorted by camera motion, different adverse weather conditions, and poor lighting conditions that significantly increase the difficulty level of this task.

TSRD is still challenging research, and several studies have been conducted on it. The most important of these studies can be found in \cite{ellahyani2021traffic}. TSRD consists of two independent tasks: Traffic Sign Detection (TSD) and Traffic Sign Recognition (TSR). Traditionally, the methods of TSD mostly depend on manual feature extraction. Features are extracted manually from various attributes such as edge detection, geometrical shapes, and color information. The color-based approach generally consists of threshold-based segmentation of traffic sign sections in a specific color space, such as Hue-Chroma-Luminance (HCL) \cite{manjare2014image}, Hue-Saturation-Intensity (HSI) \cite{huang2016efficient}, and others \cite{chen2015accurate, kashiyani2017patchwise, asgarian2021fast}. However, one major weakness of color-based approaches is that they are highly sensitive to the change in illumination, which can frequently occur in real-world scenarios \cite{temel2019traffic}. To address this challenge, shape-based approaches have been widely used in the existing literature, which includes Fast Fourier Transform (FFT) \cite{larsson2011using}, Haar-Wavelet features \cite{larsson2011correlating}, Histogram-Oriented Gradients (HOG) \cite{zhu2016traffic}, and Canny-Edge detection \cite{lu2018traffic}. Due to camera motion during real-time video feed, disorientation, scale variation, and occlusion of traffic sign regions hinder the practical application of these approaches. In contrast, TSR methodologies use Convolutional Neural Networks (CNN) to extract deep features, including MCDNN \cite{yuan2019vssa} and Robust Embedded network \cite{manzari2021robust}. Unfortunately, until today there is no comprehensive model to identify the best classifier and the best feature extraction backbone for TSR.

In this study, we particular focus on designing new backbone for traffic sign detection. As far as the authors know, no previous studies have applied Vision Transformers to this specific task. The Transformer was first introduced in the field of natural language processing (NLP), which is a deep neural network based on a self-attention mechanism. In contrast to the traditional recurrent neural networks and convolutional neural networks (CNNs), Transformer is an attention-only structure without recursion or convolution operations, which improve parallel efficiency without sacrificing the performance \cite{dosovitskiy2020image}. Vision Transformers are established as novel backbone for vision-related issues. Transformers (ViTs) have emerged with encouraging performance in computer vision tasks, and their structures are under development.

In this paper, we propose a novel Pyramid Transformer enhanced by locality inductive bias and pyramid module, which combines two kinds of basic blocks, i.e., pyramid block (PB) and normal block (NB). PBs are used to downsample and embed the input images into tokens with rich multi-scale context, while NBs attempt to jointly model locality and hierarchical global attributes in the token sequence. Furthermore, each block consists of two paralleled branches, including a paralleled local self-attention and convolutional layers followed by a feed-forward network (FFN). It is noteworthy that PB has an extra pyramid module, which comprises atrous convolutions with different dilation rates to embed local and global features into tokens. Experiments show that our scheme can effectively improve the detection speed and accuracy on GTSDB.

\begin{table*}[t]
    \caption{Performance for object detection on the GTSDB}
    \begin{adjustbox}{width=.7\textwidth,center}

        \begin{tabular}{c |c| c| c c c}
            \toprule
            Decoder & backbone & params (M) & $AP$ & $AP^{75}$ & $AP^{50}$ \\
            \midrule
            & ResNet-50 \cite{he2016deep}& 44 & 63.4 & 67.3 & 83.1 \\
            & PVT-S \cite{wang2021pyramid}& 44 & 65.4 & 67.3 & 87.7 \\
Faster RCNN & Swin-T \cite{liu2022swin}& 48 & 68.4 & 72.6 & 90.6 \\
            & DAT-T \cite{xia2022vision}& 46 & 68.3 & 72.7 & 91.1 \\
            & Ours & 36 & \teblebold{70.2} & \teblebold{74.6} & \teblebold{91.2} \\
            \midrule
            & ResNet-50 \cite{he2016deep}& 82 & 70.7 & 74.6 & 88.4\\
            & PVT-T \cite{wang2021pyramid}& 83 & 75.2 & 79.4 & 93.9\\
Cascade RCNN & Swin-T \cite{liu2022swin}& 86 & 74.5 & 78.8 & 93.3 \\
            & DAT-T \cite{xia2022vision}& 86 & 75.4 & 79.9 & 94.2 \\
            & Ours & 74 & \teblebold{77.8} & \teblebold{81.8} & \teblebold{96.5} \\
            \bottomrule
        \end{tabular}
    \end{adjustbox}
\end{table*}

\section{METHOD}

This section provides a brief overview of ViT and then describes the proposed Pyramid transformer architecture in details.

\subsection{Revisit vision transformer}

Our method is based on Vision Transformer (ViT), so we first provide a brief overview of ViT \cite{dosovitskiy2020image}. In contrast to CNN-based approaches for image classification, ViT utilizes a purely attention-based mechanism. In each transformer model, two basic components are included: a multi-head self-attention (MHSA) and a feed-forward network (FFN) with layer normalization and residual shortcuts. In order to adapt transformers to vision tasks, ViT divides images into fixed-sized patches, for example $16\times16$, and then linearly projects them into tokens. To form the input sequence, patch tokens are additionally appended with class tokens. To learn the positional information of each token, we add a learnable absolute positional embedding before feeding it to transformer encoders. At the end of the network, the class token is used as the final feature representation. ViT can be expressed as follows:

\begin{align}
   &&x_0 &= [x_{patch}||x_{cls}] + x_{pos},\\
   &&y_k &= x_{k-1} + MHSA(LN(x_{k-1})),  \\
   &&x_k &= y_k + FFN(LN(y_k)),
\end{align}

where $ x_{cls} \in R^{1\times C} $ and $ x_{patch} \in R^{N\times C} $ represent class tokens and patch tokens, respectively, while $ x_{pos} \in R^{(1+N)\times C} $ represents position embeddings, in the following equation, $K$ is the layer index, $C$ is the number of patch tokens, and $N$ is the embedding dimension. However, vanilla ViT is ideally capable of learning global interactions between all patch tokens; the memory complexity of self-attention increases when there are many tokens since the computational cost grows quadratically. As a result, the vanilla ViT model cannot extend to vision applications requiring high-resolution details, such as object detection. We propose the pyramid module for the vision transformer to address these issues.

\subsection{Overview architecture of Pyramid transformer}

Pyramid transformer aims to introduce the pyramid structure into the Vision Transformer, so that it can generate multi-scale feature maps for dense prediction tasks. As shown in Figure 1, The proposed architecture is composed of two types of blocks, PBs and NBs. The function of PBs is to embed multi-scale context into tokens, and NBs are used further to inject convolutional bias into Transformers. fig 1 shows that Pyramid Transformer offers four stages to downsample an input image of size $x \in R^{H \times W \times C}$ , where four PBs are used to reduce the features gradually by $4\times$, $2\times$, $2\times$, and $2\times$, respectively. At each stage, several normal blocks are sequentially groped following the PB. Note that all of NBs in each group have the same isotropic design. The number of normal cells determines model parameters and size. Using this method, Pyramid transformer can extract a feature pyramid from different stages ${F_1, F_2, F_3, F_4}$, which is used by decoder heads designed for different downstream tasks.

\subsection{Pyramid Block}

In each stage, pyramid blocks are employed to embed multiscale context and local information into visual tokens instead of directly splitting and flattening images using a linear image patch embedding layer, which introduces intrinsic inductive bias scale-invariance from convolutions. In Fig 1, the PB is divided into two parallel branches, which together model locality and long-range dependency, followed by an $FFN$ that transforms the features. The image $x$ is the input for the first PB, and the input feature of $i_{th} $ PB is denoted as $h_i \in R^{H_i \times W_i \times D_i}$. In the branch of the global dependency, firstly, $h_i$ is fed into a Pyramid Reduction Module (PRM) for multiscale context extraction as follows:

\begin{equation}
  PRM(f_i) = Cat([Conv_{ij}(h_i;s_{ij};r_i)])
\end{equation}

where $Conv_{ij}$ denotes the $jth$ convolutional layer in the $PRM$. The predefined dilation rate set $S_i$ is used to calculate $s_{ij}$ corresponding to the $ith$ PB. The spatial dimension of the features is reduced by a ratio $r_i$ from the predefined reduction ratio set using stride convolution. The concatenation of the features after convolution occurs along the channel dimension, namely, ${h_i}^{ms} \in R^{(W_i/r_i)\times (H_i/r_i) \times (S_i)} $, where $S$ is the number of dilation rates. Next, MHSA module then processes ${h_i}^{ms}$ to model long-range dependencies. Moreover, we embed local context through Parallel Convolutional Modules ($PCM$), which are fused as follows:

\begin{align}
  &&{h_i}^g &= MHSA_i(Img2Seq({h_i}^{ms})),\\
  &&{h_i}^{`g} &= {h_i}^g + PCM_i(h_i).
\end{align}

$PCM$ consists of sequentially stacked convolution layers, batch normalizations (BN), and an $Img2Seq$ operation. Furthermore, parallel convolutions use stride convolutions to match the $PRM$'s spatial downsampling ratio. As a result, the token features learned both low-level and multi-scale information, implying that PB acquires local and scale-invariance inductive bias by design. FFN processes the fused tokens, and then the resulting sequence is converted back to feature maps. This feature fed into the following PB or NB.

\subsection{Normal Block}

Normal blocks have a similar structure to pyramid blocks, except that there is no PRM in NBs, as shown on the right side of Fig 1. Because feature maps after PBs have relatively small spatial dimensions, $PRM$ is not needed in NBs. The class token from the third PB is concatenated with F from the third PB and then combined with positional embeddings to get the input tokens for the following NBs. Class tokens are randomly generated at the start of the training phase and fixed during the procedure. The tokens are fed into the $MHSA$ module in the same way as the PB. The resulting sequence is transformed into feature maps during this process and then fed into the $PRM$. Notably, the class token is discarded because $PRM$ has no spatial connection to other visual tokens. Stacked convolutions are used in $PRM$ to further reduce the parameters in NBs. An element-wise sum is then applied to the features from MHSA and $PRM$. Finally, the $FFN$ processes the result tokens to obtain the output features of NB.

\section{EXPERIMENTS}

In this section, as a first step, we describe the datasets used for training our Pyramid Transformer, followed by an explanation of the experimental settings. In the final step, we focus on the extensive experimental results of our Transformer model compared to SOTAs studies regarding traffic sign detection.

\subsection{Dataset and evaluation measures}\label{AA}

We test our proposed model on the German Traffic Sign Detection Benchmark (GTSDB) \cite{stallkamp2012man} dataset. Several reasons explain why this dataset is chosen over others, including its popularity and frequent use in studies to compare traffic sign detection methods. The GTSDB dataset contains natural traffic scenes recorded in weather conditions (rain, fog, sunny) and various types of roads (urban, highway, rural) captured during daylight and twilight. This dataset includes $900$ full images comprising $1206$ traffic signs, split into a train set of $600$ images ($846$ traffic signs) and a test set with $300$ images ( $360$ traffic signs). The images contain zero, one, or multiple traffic signs, which naturally suffer from differences in orientation, occlusions, or low-light conditions. The GTSDB has $43$ classes in total, also divided into four superclasses: mandatory, prohibitory, danger, and other. For the GTSDB classes, we evaluate on the subset of the $43$ traffic signs because traffic signs in the same superclass still contain different information, such as ‘‘$speed limit 30$’’ and ‘‘$speed limit 70$’’.

The mean average precision (mAP), which is commonly used as an evaluation metric in the object detection dataset \cite{lin2014microsoft}, \cite{everingham2010pascal}, is used as the standard criterion in this study. Average Precision (AP) describes the trade-off between precision and recall by describing the area under the precision-recall curve. The AP for one class object is calculated, and the mAP is the average value for all classes in the dataset. Further, the Intersection-over-Union (IoU) used in this paper has values of $0.5$ and $0.75$ for computing mAP.

\subsection{Implementation details}

We evaluate the performance of our Pyramid Transformer using two representative meta-architectures: Faster RCNN \cite{ren2015faster} and Cascade RCNN \cite{cai2018cascade}. Specifically, we use our transformer models to build the backbones of these frameworks. All the models are trained under the same setting as in \cite{liu2021swin}. We try to compare fairly and build consistent settings for future experiments. Further, we report standard schedule ($36$ epochs) detection results on the GTSDB dataset [48] in Table 1. The Faster RCNN train was performed using AdamW optimizer for $36$ epochs with $16$ batch sizes. Initially, the learning rate is $0.0001$, starting with a $600-iteration$ warmup, and declining by $0.1$ at epochs $7$ and $11$. We use stochastic drop path regularization of $0.2$ and weight decay $0.0002$. The implementation is based on MMDetection \cite{chen2019mmdetection} toolboxes. According to \cite{liu2021swin}, for $3x$ training of Cascade RCNN, we use an initial learning rate of $0.0002$ and weight decay of $0.0005$ for the whole network. All other hyperparameters follow the default settings used by Swin \cite{liu2022swin}. We follow the common multi-scale training to randomly resize the input image so that its shorter side is between $800$ and $480$ while longer side is less than $1333$.

\subsection{Comparison with the state-of-the-art}

Table 1 summarizes the results, showing that Pyramid Transformer achieves the best performance while requiring the least number of parameters. Thanks to the introduced pyramid module and locality inductive bias, With Faster RCNN as the decoder for the $36$ epochs setting, the proposed model achieved $8.1\%$ $AP^{50}$ , $7.3\%$ $AP^{75}$, and $6.8\%$ $AP$ performance gains over Swin \cite{liu2022swin}. It also considerably outperforms other backbones like ResNet and PVT, owning to our model’s efficient structure design. From table 1 our method has the highest mAP when using Cascade RCNN as the decoder, obtaining $77.8\%$ $AP$, $81.8\%$ $AP^{75}$, and $96.5\%$ $AP^{50}$ when training only 36 epochs, which our method improved $2.1\% mAP$ on Faster RCNN and $2.4\%$ $mAP$ on Cascade RCNN.  The superiority of the Pyramid Transformer is due to the fact that introducing inductive bias into the pyramid feature module helps our model better utilize the data to further improve performance for object detection.

\section{CONCLUSION}

In this work, we proposed a novel vision transformer based on two-staged frameworks for traffic sign detection, which integrates spatial structure and local information into vision Transformers using two basic blocks (Pyramid blocks and Normal blocks).  Meanwhile, the multi-scale Pyramid features from our model can be applied to various dense prediction visual tasks. Experimental evaluations on the GTSDB dataset approve that our Pyramid Transformer attains better performance compared to state-of-the-art studies. Our next steps will be to investigate more traffic signs from classes that are seldom seen in this benchmark. We also plan to enhance the speed of the process in order to run it on embedded devices in real-time.



\hyphenpenalty=10000 
\bibliographystyle{IEEEtran} 				
\bibliography{cas-refs}
\vskip8pt

\end{document}